\def\year{2020}\relax
\documentclass[letterpaper]{article} 
\usepackage{aaai20}  
\usepackage{times}  
\usepackage{helvet} 
\usepackage{courier}  
\usepackage[hyphens]{url}  
\usepackage{graphicx} 
\usepackage{graphicx}  
\usepackage{multirow}
\usepackage{booktabs}
\usepackage{float}
\usepackage{graphicx}
\usepackage{caption}
\usepackage{subcaption}
\usepackage{url}
\usepackage{array}
\usepackage{amsfonts}
\usepackage{amsmath,bm}
\usepackage{enumitem}
\usepackage[linesnumbered, ruled]{algorithm2e}
\usepackage{algorithmic}
\usepackage{pifont}

\usepackage{amssymb}
\usepackage{placeins}
\usepackage{balance}
\SetKwInput{KwData}{Input}
\SetKwInput{KwResult}{Output}
\newtheorem{problem}{Problem}

\title{Few-Shot Knowledge Graph Completion}

\author{Chuxu Zhang${}^{1}$, Huaxiu Yao${}^{2}$,  Chao Huang${}^{3}$, Meng Jiang${}^{1}$, Zhenhui Li${}^{2}$, Nitesh V. Chawla${}^{1}$\\
${}^{1}$University of Notre Dame,
${}^{2}$Pennsylvania State University,
${}^{3}$JD Finance American Corporation\\
${}^{1}$\{czhang11,mjiang2,nchawla\}@nd.edu, ${}^{2}$\{huaxiuyao,jessieli\}@psu.edu, ${}^{3}$chaohuang75@gmail.com
}

\begin{document}
\maketitle

\begin{abstract}
Knowledge graphs (KGs) serve as useful resources for various natural language processing applications.
Previous KG completion approaches require a large number of training instances (i.e., head-tail entity pairs) for every relation. The real case is that for most of the relations, very few entity pairs are available.
Existing work of one-shot learning limits method generalizability for few-shot scenarios and does not fully use the supervisory information; however, few-shot KG completion has not been well studied yet.
In this work, we propose a novel few-shot relation learning model (FSRL) that aims at discovering facts of new relations with few-shot references. FSRL can
effectively capture knowledge from heterogeneous graph structure, aggregate representations of few-shot references, and match similar entity pairs of reference set for every relation. Extensive experiments on two public datasets demonstrate that FSRL outperforms the state-of-the-art. 
\end{abstract}

\section{Introduction}
Large-scale knowledge graphs (KGs)
such as YAGO \cite{suchanek2007yago}, NELL \cite{carlson2010toward}, and Wikidata \cite{vrandevcic2014wikidata}
usually represent facts in the form of relations (edges) between (head-tail) entity pairs (nodes). This kind of graph-structured knowledge is essential for many downstream applications such as search, question answering, and semantic web. However, KGs are known for their incompleteness. In order to automate the KG completion process, many work \cite{nickel2011three,bordes2013translating,socher2013reasoning,yang2015embedding,trouillon2016complex,schlichtkrull2018modeling,dettmers2018convolutional} have been proposed to infer missing relations by learning existing ones. For example, RESCAL \cite{nickel2011three} employs tensor factorization to capture inherent structure of multi-relational data in KGs. TransE \cite{bordes2013translating} interprets relations as translation operation on the low-dimensional embeddings of entities. And recently, G-GCN \cite{schlichtkrull2018modeling} models relational structure by graph neural network.

The above methods need a good number of entity pairs for every relation. However, the frequency distributions of relations in real datasets often have long tails. A large portion of relations have only few entity pairs in KGs.
It is important and challenging to deal with the relations with limited (few-shot) number of entity pairs. The few-shot scenario incurs the infeasibility of previous models which assume available, sufficient training instances for all relations.

In light of the above issue, Xiong \textit{et al.} \shortcite{xiong2018one} proposed GMatching which introduces a local neighbor encoder to learn entity embeddings.
It achieves considerable performance in one-shot relation inference yet still has some limitations. First, GMatching assumes all local neighbors contribute equally to the entity embedding, whereas heterogeneous neighbors could have different impacts. For example, the embedding of ``Nadella'' may have more influence on the embedding of ``Microsoft'' than ``Apple'' as the company has only one CEO and a number of competitors. Thus the neighbor encoder of GMatching learns insufficient graph structure representation and impairs the model performance. Second, GMatching is designed under one-shot learning setting. Although it can be modified to few-shot case by adding a pooling layer over reference set, the general operation ignores the interaction among few-shot reference instances and limits the representation capability of reference set. Therefore, it is crucial to design a model to effectively complete relations with limited reference entity pairs.

To address the above weak points, we propose a \underline{F}ew-\underline{S}hot \underline{R}elation \underline{L}earning model (FSRL) with the purpose of learning a matching function that can effectively infer the true entity pairs given the set of few-shot reference entity pairs for each relation. To be more specific, first, we propose a relation-aware heterogeneous neighbor encoder to learn entity embeddings based on the heterogeneous graph structure and attention mechanism. It captures both different relation types and impact differences of local neighbors. Next, we design a recurrent autoencoder aggregation network to model interactions of few-shot reference entity pairs and accumulate their expression capabilities for each relation. With the aggregated embedding of reference set, we finally employ a matching network to discover similar entity pairs of reference set. The meta-training based gradient descent approach is employed to optimize model parameters. The learned model can be further applied to infer true entity pairs for any new relation without any fine-tuning step. 

To summarize, our main contributions are:
\begin{itemize}[leftmargin=*]\setlength{\itemsep}{0pt}
\item We introduce a new few-shot KG completion problem which is different from previous work and more suitable for practical scenarios. 
\item We propose a few-shot relation learning model to solve the problem. The model performs joint optimization of several learnable neural network modules.
\item We conduct extensive experiments on two public datasets. Results demonstrate that our model outperforms state-of-the-art baselines.
\end{itemize}

\section{Related Work}
Here we survey two topics relevant to this work: few-shot learning and relation learning for KGs. 

\subsection{Few-Shot Learning} 
Recent few-shot learning models have two categories: (1) metric based approaches \cite{koch2015siamese,vinyals2016matching,snell2017prototypical,mishra2018simple}; (2) meta-optimizer based approaches \cite{ravi2016optimization,finn2017model,li2017meta,finn2018probabilistic,lee2018gradient,yao2019hier}. The former one learns an effective metric and corresponding matching function among a set of training instances. For example, matching networks \cite{vinyals2016matching} make predictions by comparing the input example with a few-shot labeled support set. Prototypical networks \cite{snell2017prototypical} classify each sample by computing the distance to prototype representation of each class. The later one aims to quickly optimize the model parameters given the gradients on few-shot data instances. One example is the model-agnostic meta-learning (MAML) \cite{finn2017model} which trains model via a small number of gradient updates and leads to fast learning on a new task. Another example is the LSTM-based meta-learner \cite{ravi2016optimization} that learns the exact optimization algorithm used to train another neural network classifier in the few-shot regime. Unlike the previous few-shot learning study that focus on vision \cite{yang2018learning}, imitation learning  \cite{duan2017one}, spatiotemporal analysis~\cite{yao2019learning}, sentiment analysis~\cite{li2019transferable} domains, we leverage few-shot learning to complete KGs.

\subsection{Relation Learning for KGs} 
Many work have been proposed to model relational structure in KGs and automate KG completion. For example, Nickel \textit{et al.} \shortcite{nickel2011three} designed RESCAL to model inherent structure of dyadic relational data by tensor factorization. Bordes \textit{et al.} \shortcite{bordes2013translating} proposed TransE that interprets relationships as translation operating on the low-dimensional embeddings of the entities. Unlike representing entities with single vectors, Socher \textit{et al.} \shortcite{socher2013reasoning} developed NTN that represents entities as an average of their constituting word vectors. Later, more sophisticated models have been proposed, such as DistMul \cite{yang2015embedding} and ComplEx \cite{trouillon2016complex}. Recently, deep neural network based models like R-GCG \cite{schlichtkrull2018modeling} and ConvE \cite{dettmers2018convolutional} have been presented for further improvement. Different from those models that assume sufficient training instances are available,
Xiong \textit{et al.} \shortcite{xiong2018one} presented GMatching model for one-shot relation learning in KGs. In this work, we study a practical few-shot scenario which deals with long tail or newly added relations with few-shot reference instances. 

\section{Preliminaries}
In this section, we formally define the few-shot knowledge graph completion problem and detail the corresponding few-shot learning settings.

\subsection{Problem Definition}
A KG $G$ is represented as a collection of triples $\{(h, r, t)\} \subseteq  \mathcal{E} \times \mathcal{R}  \times \mathcal{E}$, where $\mathcal{E}$ and $\mathcal{R}$ denote the entity set and relation set, respectively. The KG completion task is to either predict the tail entity $t$ given the head entity $h$ and the query relation $r$: $(h, r, ?)$, or predict unseen relation $r$ between head entity and tail entity: $(h, ?, t)$. In this work, we focus on the former case as we want to predict the unseen facts of a given relation. Unlike previous studies that assume enough entity pairs 
are available for each relation, this work considers a practical scenario that few-shot entity pairs (reference set) are given. The purpose is to rank the true entity $t_{true}$ higher than false candidate entities $t_{false}$, given few-shot reference pairs $(h_{k}, t_{k}) \in R_{r}$ of relation $r$. Formally, the problem is defined as follows:
\begin{problem}
{\bf Few-Shot Knowledge Graph Completion} Given the relation $r$ and its few-shot reference entity pairs $(h_{k}, t_{k}) \in R_{r}$, the task is to design a machine learning model which ranks all tail candidate entities $t$ for each new head entity $h$, such that the top ranked $t$ are true tail entities of $h$. 
\end{problem}
The candidate entities set is constructed based on the entity type constraint \cite{xiong2018one}, and we only consider a closed set of entities which excludes the unseen entities when predicting facts of new relations in test period. 

\subsection{Few-Shot Learning Settings}
The purpose of this work is to design a machine learning model which could be utilized to predict the new facts with few-shot reference instances. Following the standard few-shot learning settings \cite{ravi2016optimization,snell2017prototypical}, we can access to a set of training tasks. In the problem, each training task corresponds to a KG relation $r \in \mathcal{R}$ with its own training/testing entity pairs data: $D_{r} = \left \{P^{train}_{r}, P^{test}_{r}  \right \}$. We denote this kind of task set as meta-training set, $\mathcal{T}_{mtr}$. To imitate the few-shot relation prediction in evaluation period, each $P^{train}_{r}$ only contains few-shot entity pairs $(h_{k}, t_{k}) \in R_{r}$. Besides, $P^{test}_{r} = \left \{ (h_{i},t_{i}, C_{h_{i},r}) | (h_{i},r, t_{i}) \in G\right \}$ contains all testing entity pairs of $r$, including true tail entities $t_{i}$ of each query $(h_{i}, r)$ and the remaining candidate entities $t_{j} \in C_{h_{i},r}$ where $t_{j}$ is an entity in $G$. The proposed model thus could be tested on this set by ranking all candidate entities given the test query $(h_{i}, r)$ and the few-shot reference pairs in $P^{train}_{r}$. We denote the ranking loss of relation $r$ as $\mathcal{L}_{\Theta}(h_{i}, t_{i}|C_{h_i, r}, P^{train}_{r})$, where ${\Theta}$ is the set of model parameters.
Thus, the objective of model training is defined as:
\begin{equation}
\begin{gathered}
min_{\Theta}\mathbb{E}_{\mathcal{T}_{mtr}} \left [\sum_{(h_{i}, t_{i}, C_{h_{i}, r}) \in P^{test}_{r}} \frac{\mathcal{L}_{\Theta}(h_{i}, t_{i}|C_{h_i, r}, P^{train}_{r})}{|P^{test}_{r}|} \right ]
\end{gathered}
\label{equ: objective-meta-train}
\end{equation} 
where 
$|P^{test}_{r}|$ represents the number of tuples in $P^{test}_{r}$. In next section, we will detail how to formulate and optimize the above objective function.

After sufficient training, the learned model can be utilized to predict facts of each new relation $r' \in\mathcal{R}'$. This step is called the meta-testing. The relations in meta-testing are unseen from meta-training, i.e., $\mathcal{R} \cap \mathcal{R}' = \phi$. The same as meta-training relations, each relation $r'$ in meta-testing has its own few-shot training data $P^{train}_{r'}$ and testing data $P^{test}_{r'}$. These relations form a meta-testing set which is denoted as $\mathcal{T}_{mte}$. 
In addition, we leave out a subset of relations in $\mathcal{T}_{mtr}$ as the meta-validation set $\mathcal{T}_{mtv}$. Furthermore, the model can access to a background KG $G'$, which is a subset of $G$ that excludes all the relations in $\mathcal{T}_{mtr}$, $\mathcal{T}_{mte}$ and $\mathcal{T}_{mtv}$. 

\section{Model}
\begin{figure*}[hbt!]
\begin{center}
\includegraphics[scale=0.71]{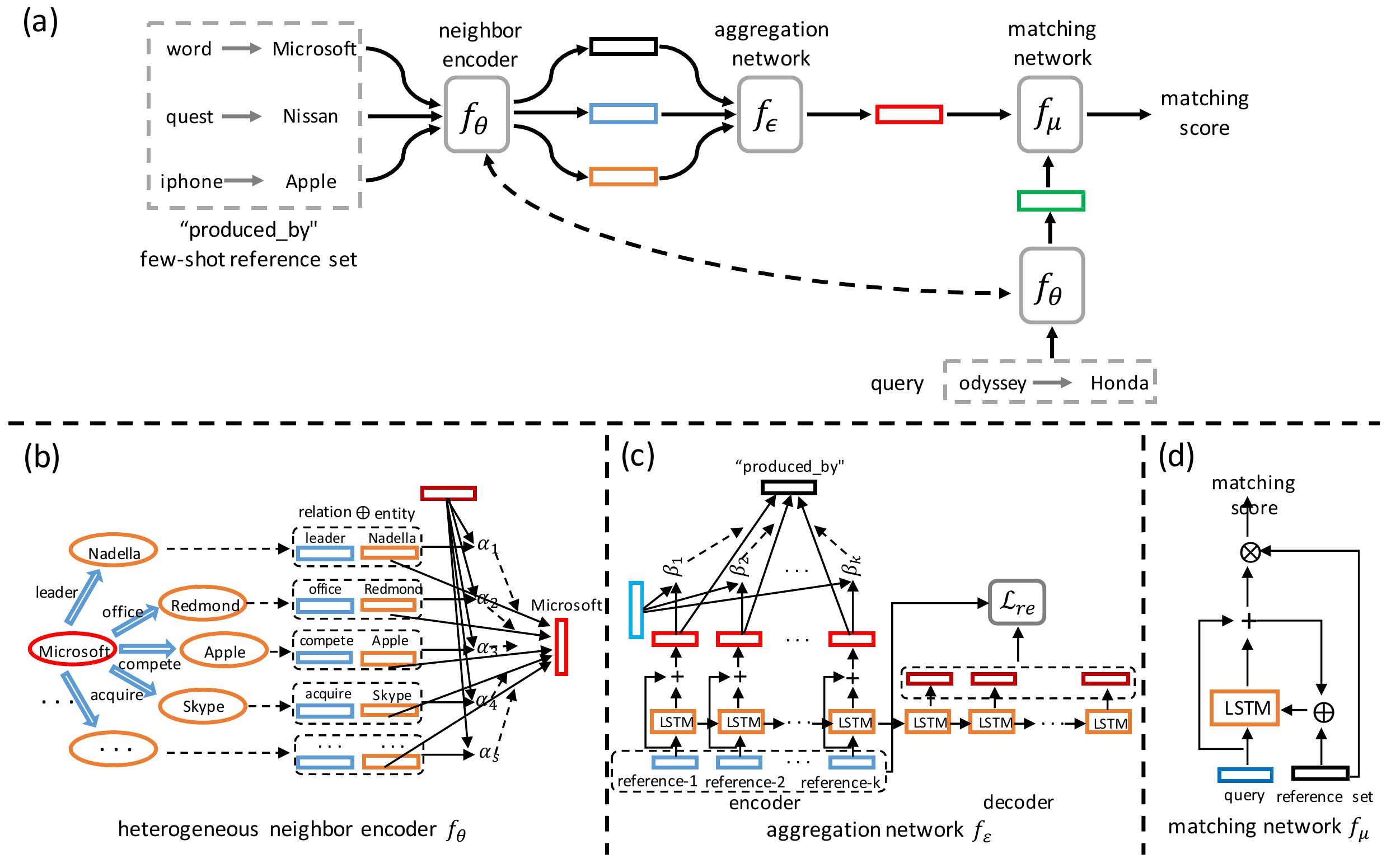}
\caption{(a) The framework of FSRL: it first generates entity embedding via heterogeneous neighbor encoder, then aggregates few-shot reference entity pairs and generate reference set embedding, finally employs a matching network to compute similarity score between query pair and reference set; (b) the relation-aware heterogeneous neighbor encoder for entity; (c) the recurrent autoencoder aggregation network for reference set; (d) the recurrent matching network for query pair and reference set.} 
\label{fig: model}
\end{center}
\end{figure*}

In this section, we present the detail of FSRL. FSRL consists of three major parts: (1) encoding heterogeneous neighbors for each entity; (2) aggregating few-shot reference entity pairs for each relation; (3) matching query pairs with reference set for relation prediction. Figure \ref{fig: model} shows the framework of FSRL.  

\subsection{Encoding Heterogeneous Neighbors}
Although many work \cite{nickel2011three,bordes2013translating,yang2015embedding} have been proposed to learn entity embeddings by using relational information, Xiong \textit{et al.} \cite{xiong2018one} demonstrated that explicitly encoding graph local structure (i.e., one-hop neighbors) can benefit relation prediction. The proposed neighbor encoder takes the average of feature representations of all relational neighbors as the embedding of given entity. Despite the desirable performance, it neglects the different impacts of heterogeneous neighbors which may help improve entity embedding \cite{zhang2019hetgnn}. In light of this issue, we design a relation-aware heterogeneous neighbor encoder. 
Specifically, we denote the set of relational neighbors ($relation, entity$) of given head entity $h$ as $\mathcal{N}_{h} = \left \{ (r_{i}, t_{i}) | (h, r_{i}, t_{i})\in G'\right \}$, where $G'$ is the background knowledge graph, $r_{i}$ and $t_{i}$ represent the $i$-th relation and corresponding tail entity of $h$, respectively. The heterogeneous neighbor encoder should be able to encode $\mathcal{N}_{h}$ and output a feature representation of $h$ by considering different impacts of relational neighbors $(r_{i}, t_{i}) \in \mathcal{N}_{h}$. To achieve this goal, we introduce an attention module and formulate the embedding of $h$ as follows: 
\begin{equation}
\begin{gathered}
f_{\theta}(h) = \sigma \left. \big( \sum _{i}\alpha_{i} e_{t_{i}}\right. \big)\\
\alpha_{i} = \frac{exp\left. \big\{u^{T}_{rt} \left. \big(\mathcal{W}_{rt}(e_{r_{i}}\oplus e_{t_{i}}) + b_{rt}\right. \big)\right. \big\}}{\sum_{j}exp\left. \big\{u^{T}_{rt}\left. \big(\mathcal{W}_{rt}(e_{r_{j}}\oplus e_{t_{j}}) + b_{rt}\right. \big)\right. \big\}} 
\end{gathered}
\label{equ: neighbor_encoder_1}
\end{equation} 
where $\sigma$ denotes activation unit (we use Tanh), $\oplus$ represents concatenation operator, $e_{t_{i}}$, $e_{r_{i}} \in \mathbb{R}^{d \times 1}$ are pre-trained embeddings of $t_{i}$ and $r_{i}$. Besides, $u_{rt} \in \mathbb{R}^{d \times 1}$, $\mathcal{W}_{rt} \in \mathbb{R}^{d \times 2d}$ and $b_{rt} \in \mathbb{R}^{d \times 1}$ ($d$: pre-trained embedding dimension) are learnable parameters. Figure \ref{fig: model}(b) illustrates the detail of heterogeneous neighbor encoder.  
According to Eq. \ref{equ: neighbor_encoder_1}, the formulation of $f_{\theta}(h)$ considers the different impacts of heterogeneous relational neighbors via attention weight $\alpha_{i}$ and leverages both embeddings of entity $t_{i}$ and relation $r_{i}$ to compute $\alpha_{i}$.

\subsection{Aggregating Few-Shot Reference Set}
The current models (e.g., GMatching) are not able to model the interactions of few-shot instances in reference set, which limits model capability.
Thus, we need to design a module to effectively formulate the aggregated embedding of reference set $R_{r}$ for each relation $r$. By applying the neighbor encoder $f_{\theta}(h)$ to each entity pair $(h_{k}, t_{k}) \in R_{r}$, we can obtain the representation of $(h_{k}, t_{k})$ in the form $\mathcal{E}_{h_{k}, t_{k}} = [f_{\theta}(h_{k})\oplus  f_{\theta}(t_{k})]$. Learning the representation of a reference set $R_{r}$ with few-shot entity pairs is challenging
as it requires modeling interactions among different entity pairs and accumulating their expression capability. 
Inspired by the common practices in learning sentence embeddings~\cite{conneau2017supervised} in natural language processing and aggregating node embeddings~\cite{hamilton2017inductive} in graph neural networks, we tackle the challenge and formulate the embedding of $R_{r}$ by aggregating representations of all entity pairs in $R_{r}$:
\begin{equation}
\begin{split}
f_{\epsilon }(R_{r}) = \mathcal{AG}_{(h_{k}, t_{k}) \in R_{r}}\left. \big\{ \mathcal{E}_{h_{k}, t_{k}} \right. \big\}
\end{split}
\label{equ: neigh_aggregator}
\end{equation} 
where $\mathcal{AG}$ is an aggregation function which can be pooling operation, feed-forward neural network, etc. Motivated by the recent success of recurrent neural network aggregator in order-invariant problems such as graph embedding~\cite{hamilton2017inductive}, we design a recurrent autoencoder aggregator which achieves good capability. Specifically, the entity pair embeddings $\mathcal{E}_{h_{k}, t_{k}} \in R_{r}$ are sequentially fed into a recurrent autoencoder by: 
\begin{equation}
\begin{gathered}
\mathcal{E}_{h_{1}, t_{1}}\rightarrow m_{1}\rightarrow \cdots \rightarrow  m_{K} \rightarrow d_{K} \rightarrow \cdots \rightarrow d_{1}
\end{gathered}
\label{equ: neigh_aggregator_1}
\end{equation} 
where $K$ is the size of reference set (i.e., few-shot size). The hidden states $m_{k}$ and $d_{k}$ of encoder and decoder are computed by: 
\begin{equation}
\begin{gathered}
m_{k} = \mathrm{RNN}_{encoder}(\mathcal{E}_{h_{k}, t_{k}}, m_{k-1}) \\
d_{k-1} = \mathrm{RNN}_{decoder}(d_{k})
\end{gathered}
\label{equ: neigh_aggregator_2}
\end{equation} 
where $\mathrm{RNN}_{encoder}$ and $\mathrm{RNN}_{decoder}$ represent recurrent encoder and decoder (e.g., LSTM \cite{hochreiter1997long}), respectively. The reconstruction loss for optimizing autoencoder is defined as:
\begin{equation}
\begin{gathered}
\mathcal{L}_{re} (R_{r}) = \sum_{k} \left \| d_{k} - \mathcal{E}_{h_{k}, t_{k}} \right \|^{2}_{2} 
\end{gathered}
\label{equ: autoencoder_loss}
\end{equation} 
$\mathcal{L}_{re}$ will be incorporated to the relational ranking loss for refining the representation of each entity pair, as we will describe later. In order to formulate the embedding of reference set, we aggregate all hidden states of encoder and extend them by adding residual connection \cite{he2016deep} and attention weight. Formally, $f_{\epsilon }(R_{r})$ is computed by:
\begin{equation}
\begin{gathered}
m'_{k} = m_{k} + \mathcal{E}_{h_{k}, t_{k}}\\
\beta_{k} = \frac{exp\left. \big\{u^{T}_{R} \left. \big(\mathcal{W}_{R}m'_{k} + b_{R} \right. \big)\right. \big\}}{\sum_{k'}exp\left. \big\{u^{T}_{R} \left. \big(\mathcal{W}_{R}m'_{k'} + b_{R} \right. \big)\right. \big\}} \\
f_{\epsilon }(R_{r}) = \sum _{k}\beta_{k} m'_{k}
\end{gathered}
\label{equ: neigh_aggregator}
\end{equation} 
where $u_{R} \in \mathbb{R}^{d \times 1}$, $\mathcal{W}_{R} \in \mathbb{R}^{d \times 2d}$ and $b_{R} \in \mathbb{R}^{d \times 1}$ ($d$: aggregated embedding dimension) are learnable parameters. Figure \ref{fig: model}(c) illustrates the detail of recurrent autoencoder aggregator. The formulation of $f_{\epsilon }(R_{r})$ aggregates all representations of $\mathcal{E}_{h_{k}, t_{k}} \in R_{r}$ and each component in this module will make effect for better performance, as we will show in the ablation study experiment. 

\subsection{Matching Query and Reference Set}
With the heterogeneous neighbor encoder $f_{\theta}$ and the reference set aggregator $f_{\epsilon}$, we now present how to effectively match each query entity pair $(h_{l}, t_{l}) \in Q_{r}$ ($Q_{r}$ is the set of all query pairs of relation $r$) with the reference set $R_{r}$. By applying $f_{\theta}$ and $f_{\epsilon}$ to the query entity pair $(h_{l}, t_{l})$ and the reference set $R_{r}$, we can obtain two embedding vectors $\mathcal{E}_{h_{l}, t_{l}} = [f_{\theta}(h_{l})\oplus  f_{\theta}(t_{l})]$ and $f_{\epsilon}(R_{r})$, respectively. In order to measure the similarity between two vectors, we employ a recurrent processor \cite{vinyals2016matching} $f_{\mu}$ to perform multiple steps matching. The $t$-th process step is formulated as:
\begin{equation}
\begin{gathered}
g'_{t}, c_{t} = \mathrm{RNN}_{match}(\mathcal{E}_{h_{l}, t_{l}}, [g_{t-1}\oplus f_{\epsilon}(R_{r})], c_{t-1}) \\
g_{t} = g'_{t} + \mathcal{E}_{h_{l}, t_{l}}
\end{gathered}
\label{equ: matching_net}
\end{equation} 
where $\mathrm{RNN}_{match}$ is LSTM cell \cite{hochreiter1997long} with input $\mathcal{E}_{h_{l}, t_{l}}$, hidden state $g_{t}$ and cell state $c_{t}$. The last hidden state $g_{T}$ after $T$ ``processing'' step is the refined embedding of query pair $(h_{l}, t_{l})$: $\mathcal{E}_{h_{l}, t_{l}} = g_{T}$. We use the inner product between $\mathcal{E}_{h_{l}, t_{l}}$ and $f_{\epsilon}(R_{r})$ as the similarity score for later ranking optimization procedure. Figure \ref{fig: model}(d) shows the detail of matching processor. This module is effective for improving model performance, as we will demonstrate in the ablation study experiment. 

\subsection{Objective and Model Training}
For the query relation $r$, we randomly sample a set of few positive (true) entity pairs $\{(h_{k}, t_{k})|(h_{k}, r , t_{k}) \in G\}$ and regard them as the reference set $R_{r}$. The remaining positive entity pairs $\mathcal{PE}_{r} = \{(h_{l}, t_{l})|(h_{l}, r, t_{l}) \in G \cap (h_{l}, t_{l}) \notin R_{r}\}$ are utilized as positive query pairs. Besides, we construct a group of negative (false) entity pairs $\mathcal{NE}_{r} = \{(h_{l}, t^{-}_{l})|(h_{l}, r, t^{-}_{l}) \notin G\}$ by polluting the tail entities. Therefore the ranking loss is formulated as:
\begin{equation}
\begin{gathered}
\mathcal{L}_{rank} = \sum _{r}\sum _{(h_{l}, t_{l}) \in \mathcal{PE}_{r}} \sum _{(h_{l}, t^{-}_{l}) \in \mathcal{NE}_{r}} \Big[\xi +s_{(h_{l}, t^{-}_{l})}-s_{(h_{l}, t_{l})}\Big]_{+} 
\end{gathered}
\label{equ: loss_rank}
\end{equation} 
where $[x]_{+} = max[0, x]$ is standard hinge loss and $\xi$ is safety margin distance, $s_{(h_{l}, t_{l})}$ and $s_{(h_{l}, t^{-}_{l})}$ are similarity scores between query pairs $(h_{l}, t_{l}/t^{-}_{l})$ and reference set $R_{r}$. By leveraging the reconstruction loss $\mathcal{L}_{re}$ of reference set aggregator, we define the final objective function as:
\begin{equation}
\begin{gathered}
\mathcal{L}_{joint} = \mathcal{L}_{rank} + \gamma  \mathcal{L}_{re}
\end{gathered}
\label{equ: loss_joint}
\end{equation} 
where $\gamma$ is trade-off factor between $\mathcal{L}_{rank}$ and $\mathcal{L}_{re}$. To minimize $\mathcal{L}_{joint}$ and optimize model parameters, we take each relation as a task and design a batch sampling based meta-training procedure. The detail of this process is summarized in Algorithm \ref{alg: IRL4BG}. 

\begin{algorithm}[tp]
\caption{FSRL Meta-Training}\label{alg: IRL4BG}
\SetKwInOut{Input}{input}
\SetKwInOut{Output}{output}
\small
\Input{Meta-training task (relation) set $\mathcal{T}_{mtr}$\\ Pre-trained KG embeddings \\
Initial model parameters $\theta$, $\epsilon$ and $\mu $}
\While{not done}{
Shuffle tasks (relations) in $\mathcal{T}_{mtr}$\\
\For {$\mathcal{T}_{r} \in \mathcal{T}_{mtr}$}{
Sample few-shot entity pairs as reference set \\
Sample a batch of query entity pairs $(h_{l}, t_{l})$\\
Pollute the tail entity of $(h_{l}, t_{l})$ to get $(h_{l}, t^{-}_{l})$ \\
Accumulate the loss by Eq. \ref{equ: loss_joint}\\
Update parameters by Adam optimizer\\
}
}
\Return Optimal model parameters $\theta^{*}$, $\epsilon^{*}$ and $\mu^{*}$ \\
\end{algorithm}

\section{Experiments}
In this section, we conduct extensive experiments to evaluate the performance of proposed model and verify the effectiveness of each component in the model. Few-shot size impact analysis and embedding visualization are also provided. 

\subsection{Experimental Design}

\subsubsection{Datasets}
We use two public datasets for experiments. The first one is based on NELL \cite{mitchell2018never}, a system that continuously collects structured knowledge from webs. The second one is based on Wikidata \cite{vrandevcic2014wikidata}. Table \ref{tab: data} lists the statistics of two datasets. The same as \cite{xiong2018one}, we select the relations with less than 500 but more than 50 triples as few-shot tasks. There are 67 and 183 tasks in NELL and Wiki data, respectively. In addition, we use 51/5/11 task relations for training/validation/testing in NELL and the division is set to 133/16/34 in Wiki.  

\begin{table}[hbt!]
\caption{Statistics of datasets. \# Ent. denotes the number of all unique entities and \# triples denotes the number of all relational triples. \# Rel. represents the number of all relations and \# Tasks represents the number of relations selected as few-shot tasks.}
\begin{center}
\small
\begin{tabular}{c||c|c|c|c}
  \toprule
  Dataset& \# Ent. & \# Triples & \# Rel. & \# Tasks \\
   \midrule
    NELL & 68,545 & 181,109 & 358  & 67\\
  Wiki & 4,838,244 & 5,859,240  & 822 & 183\\
    \midrule
  \end{tabular}
\end{center}
\label{tab: data}
\end{table}%

\subsubsection{Baseline Methods}
We consider two categories of baseline methods for comparison:
\begin{itemize}[leftmargin=*]\setlength{\itemsep}{0pt}
\item \textbf{Relational embedding methods.} This type of model learns entity/relation embeddings by modeling relational structure in KG. We employ four widely used methods: RESCAL \cite{nickel2011three}, TransE \cite{bordes2013translating}, DistMul \cite{yang2015embedding}, and ComplEx \cite{trouillon2016complex}. All entity pairs of background relations and training relations, as well as few-shot training entity pairs of validate and test relations are used to train models. 
\item \textbf{Graph neighbor encoder methods.} This type of model joints graph local neighbor encoder and matching network to learn entity embeddings and predict facts of new relations. We employ state-of-the-art model GMatching \cite{xiong2018one} for comparison. Note that there are few-shot embeddings of entity pairs in reference set, we use max/mean pooling (denoted as MaxP and MeanP) to obtain the general embedding of reference set. Moreover, we also consider taking the maximum of similarity scores between a query and all $K$ references as the final ranking score of this query. Thus in total, this type of model includes three baseline methods which are denoted as GMatching (MaxP), GMatching (MeanP), and GMatching (Max).
\end{itemize}

\begin{table*}[hbt!]
\caption{The overall results of all methods. GMatching is the best baseline. Our model has the best performances in all cases.} 
\begin{center}
\small
\begin{tabular}{l||c|c|c|c|c|c|c|c}
\toprule
 & \multicolumn{4}{c|}{Data: NELL}  & \multicolumn{4}{c}{Data: Wiki}\\
  \cmidrule{2-9}  
 Model&Hits@1 & Hits@5&Hits@10 &MRR & Hits@1 & Hits@5&Hits@10 &MRR\\
  \midrule
  \midrule
  RESCAL & .069/.141 & .160/.313 &.204/.383 &.119/.223 & .259/.057 & .297/.090 &.309/.126  & .279/.081 \\
  TransE & .056/.119 & .112/.256&.189/.320 & .104/.193  &.186/.069  &.352/.134 &.431/.176  &.273/.111 \\
  DistMult & .066/.164 & .123/.306& .178/.375 & .109/.231 & .271/.069 & \underline{.419}/.156 &  .459/.195& .339/.112\\
  ComplEx &.049/.129  & .092/.223&.112/.273 & .079/.185 &.226/.085  & .315/.117&.397/.145  &.282/.106 \\
   \midrule
   GMatching (MaxP) & .244/\underline{.198} & .418/\underline{.370} &.524/\underline{.464} & .331/\underline{.279} &\underline{.313}/.095  & .402/.235 & .468/.324 &.346/.171 \\
  GMatching (MeanP)  & \underline{.257}/.186& \underline{.455}/.360 & \underline{.542}/.453 & \underline{.341}/.267 & .290/.128 & 407/.274 & \underline{.484}/.350 & \underline{.352}/.203\\
GMatching (Max)  & .179/.152 & .391/.335 &.476/.445 & .273/.241 & .279/\underline{.135} &.396/\underline{.284} & .477/\underline{.374} & .342/\underline{.214}\\
 \midrule
 FSRL (Ours) & \textbf{.345/.211} & \textbf{.502/.433} & \textbf{.570/.507} & \textbf{.421/.318} &\textbf{.338/.155} & \textbf{.430/.327} & \textbf{.486/.406} & \textbf{.390/.241} \\
 \midrule
\end{tabular}
\end{center}
\label{tab: result}
\end{table*}%
\subsubsection{Reproducibility Settings}
The above relational embedding methods can be utilized to pre-train KG embeddings, which are further used as the input for GMatching and FSRL. We select ComplEx for pre-training as GMatching and FSRL with it achieve best performances in most cases. 
For the proposed model, we tune hyper-parameters on the validation dataset. The embedding dimension is set to 100 and 50 for NELL and Wiki dataset, respectively. The maximum number of local neighbors in heterogeneous neighbor encoder is set to 30 for both datasets. In addition, we use LSTM as the reference set aggregator and matching processor. The dimension of LSTM's hidden state is set to 200 and 100 for NELL and Wiki dataset, respectively. The number of recurrent steps equals 2 in matching network. We use the Adam optimizer \cite{kingma2014adam} to update model parameters. The initial learning rate equals 0.001 and the weight decay is 0.25 for each 10k training steps. The margin distance and trade-off factor in the objective function are set to 5.0 and 0.0001, respectively. In entity candidate set construction, we set the maximum size to 1000 for both datasets. We employ Pytorch\footnote{https://pytorch.org/} to implement our model and further conduct it on a server with GPU machines.

\subsubsection{Evaluation Metrics}
Relations and their entity pairs in training data are utilized to train the model while those of validation and test data are respectively used to tune and evaluate model. We use the top-k hit ratio (Hits@k) and the mean reciprocal rank (MRR) to evaluate performances of different methods. The k is set to 1, 5, and 10. The few-shot size $K$ is set to 3 for the following experiments. In addition, we also conduct experiment to analyze the impact of $K$. 

\subsection{Results Comparison}

\subsubsection{Overall Comparison with Baselines}
The performances of all models are reported in Table \ref{tab: result}, where the best results are highlighted in bold and the best baseline results are indicated by underline. The former/later score denotes result in validation/test dataset. According to this table: 
\begin{itemize}[leftmargin=*]
\item The graph neighbor encoder methods (GMatching) outperform the relational embedding methods, showing that incorporating graph local structure and matching network is effective for learning entity embeddings and predicting facts of new relations. 
\item FSRL achieves the best performances in all cases. The average relative improvement (\%) over the best baseline method is up to 34\% and 15\% in NELL and Wiki data, respectively. It demonstrates the effectiveness of our model. The heterogeneous neighbor encoder and recurrent autoencoder aggregation network benefit few-shot relation prediction in KGs. 
\end{itemize}

\subsubsection{Comparison Over Different Relations}
Besides the overall performance for all relations, we also conduct experiments to evaluate model performance for each relation in NELL test data. Table \ref{tab: result_relation} reports the results of FSRL and GMatching. The better result for each case is highlighted in bold. According to this table:
\begin{itemize}[leftmargin=*]
\item The results of both models on different relations are of high variance. It is reasonable since different relations have different sizes of candidate set for evaluation. The relations (e.g., relation 1) with small candidate set (or are easily to be predicted) have relative large scores and two models in these cases are compared. 
\item FSRL has better performances than GMatching in most cases. It demonstrates that our model is robust for different relations and outperforms GMatching for most relations. 
\end{itemize}

\begin{table}[hbt!]
\caption{The results of GMatching and FSRL for each relation (RId) in NELL test data.} 
\begin{center}
\small
\begin{tabular}{c||c|c|c|c|c}
\toprule
 RId& Model &Hits@1 & Hits@5&Hits@10 &MRR\\
  \midrule
  \midrule
\multirow{2}{*}{1}& GMatching & 0.946&\textbf{1.000} &\textbf{1.000} & 0.970  \\
& FSRL  &\textbf{0.972} & 0.986& \textbf{1.000}& \textbf{0.981}  \\
  \midrule
\multirow{ 2}{*}{2}& GMatching &0.279 & 0.390&0.451 & 0.360\\
&  FSRL&\textbf{0.972} & \textbf{0.972}&\textbf{0.986} &\textbf{0.975}   \\
  \midrule
\multirow{ 2}{*}{3}& GMatching & 0.069 &0.115 &0.152 & 0.108  \\
& FSRL& \textbf{0.466}& \textbf{0.418}& \textbf{0.357}&  \textbf{0.398}\\
  \midrule
\multirow{ 2}{*}{4}& GMatching&0.017 &0.033 &0.070 &0.035  \\
& FSRL  &\textbf{0.044} &\textbf{0.215} & \textbf{0.343}& \textbf{0.132}  \\
  \midrule
\multirow{ 2}{*}{5}& GMatching&\textbf{0.069} & 0.115&0.151 & 0.107 \\
&  FSRL & 0.044&\textbf{0.215} & \textbf{0.343}& \textbf{0.132}  \\
  \midrule
\multirow{ 2}{*}{6}&GMatching&0.192 &0.515 &0.581 & 0.338 \\
&  FSRL & \textbf{0.342}& \textbf{0.549}& \textbf{0.617}& \textbf{0.442}  \\
  \midrule
\multirow{ 2}{*}{7}& FSRL& \textbf{0.478}& 0.692& 0.804& \textbf{0.575} \\
& GMatching &0.438& \textbf{0.716}&\textbf{0.842} &0.562  \\
  \midrule
\multirow{ 2}{*}{8}& GMatching&0.151 &0.502  &0.669&0.312  \\
& FSRL  & \textbf{0.201}& \textbf{0.543}&\textbf{0.681} &\textbf{0.347}   \\
  \midrule
\multirow{ 2}{*}{9}&GMatching &\textbf{0.449} &0.707 &0.737&\textbf{0.564} \\
& FSRL &0.163 & \textbf{0.750}&\textbf{0.838} &  0.408 \\
  \midrule
\multirow{ 2}{*}{10}&GMatching &0.043 & 0.129&0.206 & 0.098 \\
& FSRL  &\textbf{0.069} &  \textbf{0.192}&\textbf{0.291}&\textbf{0.139}  \\
  \midrule
\multirow{ 2}{*}{11}&GMatching & 0.076& \textbf{0.708} &0.736&0.341\\
&FSRL  & \textbf{0.104}&0.631 & \textbf{0.781}& \textbf{0.416}  \\
 \midrule
\end{tabular}
\end{center}
\label{tab: result_relation}
\end{table}%

\subsection{Ablation Study}
FSRL is a joint learning framework of several neural network modules. To investigate the contributions of different components, we conduct the following ablation studies from three perspectives in Table~\ref{tab:ablation_study}, where the results in NELL data are reported and best results are highlighted in bold:
\begin{itemize}[leftmargin=*]
\item {\bf(AS\_{1})} We investigate the effectiveness of relation-aware heterogeneous neighbor encoder. We replace it by a mean pooling layer over all neighbors' embeddings. As shown in the table, our model has much better performances than the variant (in AS\_{1}), indicating the large benefit of heterogeneous neighbor encoder.
\item {\bf (AS\_{2})} We analyze the impacts of different modules of aggregation network in (AS\_{2a})-(AS\_{2c}). 
In (AS\_{2a}), we replace the recurrent autoencoder aggregation with mean pooling operation.
In (AS\_{2b}), we replace the attention weight of recurrent autoencoder with a mean pooling layer.
In (AS\_{2c}), we remove the decoder part, and only use encoder for aggregation.
According to the results in table, our model outperforms all of three variants in most cases, demonstrating the effect of each component in aggregation network.  
\item {\bf (AS\_{3})} We further analyze the effectiveness of matching network. We remove the LSTM cell and use inner-product between query embedding and reference embedding as similarity (ranking) score. 
From the results in table, our model largely outperforms the variant (in AS\_{3}), showing that recurrent matching network has good capability in computing relevance between query and reference. 
\end{itemize}

\begin{table}[hbt!]
\caption{Results of model variants in NELL data. Our model has better performance than all model variants.} 
\begin{center}
\small
\begin{tabular}{l||c|c|c|c}
\toprule
 Model&Hits@1 & Hits@5&Hits@10 &MRR  \\
  \midrule
  \midrule
  (AS\_{1})  & .121/.179 & .312/.379&.432/.464 &.212/.272   \\\midrule
  (AS\_{2a})& .281/.191& .463/.414 &.538/.504 & .368/.297  \\ 
(AS\_{2b}) &.298/.219  &.480/.420 &.556/.498 & .382/.315   \\
(AS\_{2c}) &.333/\textbf{.226}  &.478/.418 &.552/.499 & .401/.313   \\
\midrule
  (AS\_{3})& .282/.205 &.463/.394 &.548/.478 &.370/.299  \\
  \midrule
  Ours &\textbf{.345}/.211 & \textbf{.502/.433} & \textbf{.570/.507} & \textbf{.421/.318}\\
  \midrule
\end{tabular}
\end{center}
\label{tab:ablation_study}
\end{table}%

\subsection{Analysis}
\subsubsection{Impact of Few-Shot Size}
\begin{figure}[hbt!]
\begin{center}
\includegraphics[scale=0.255]{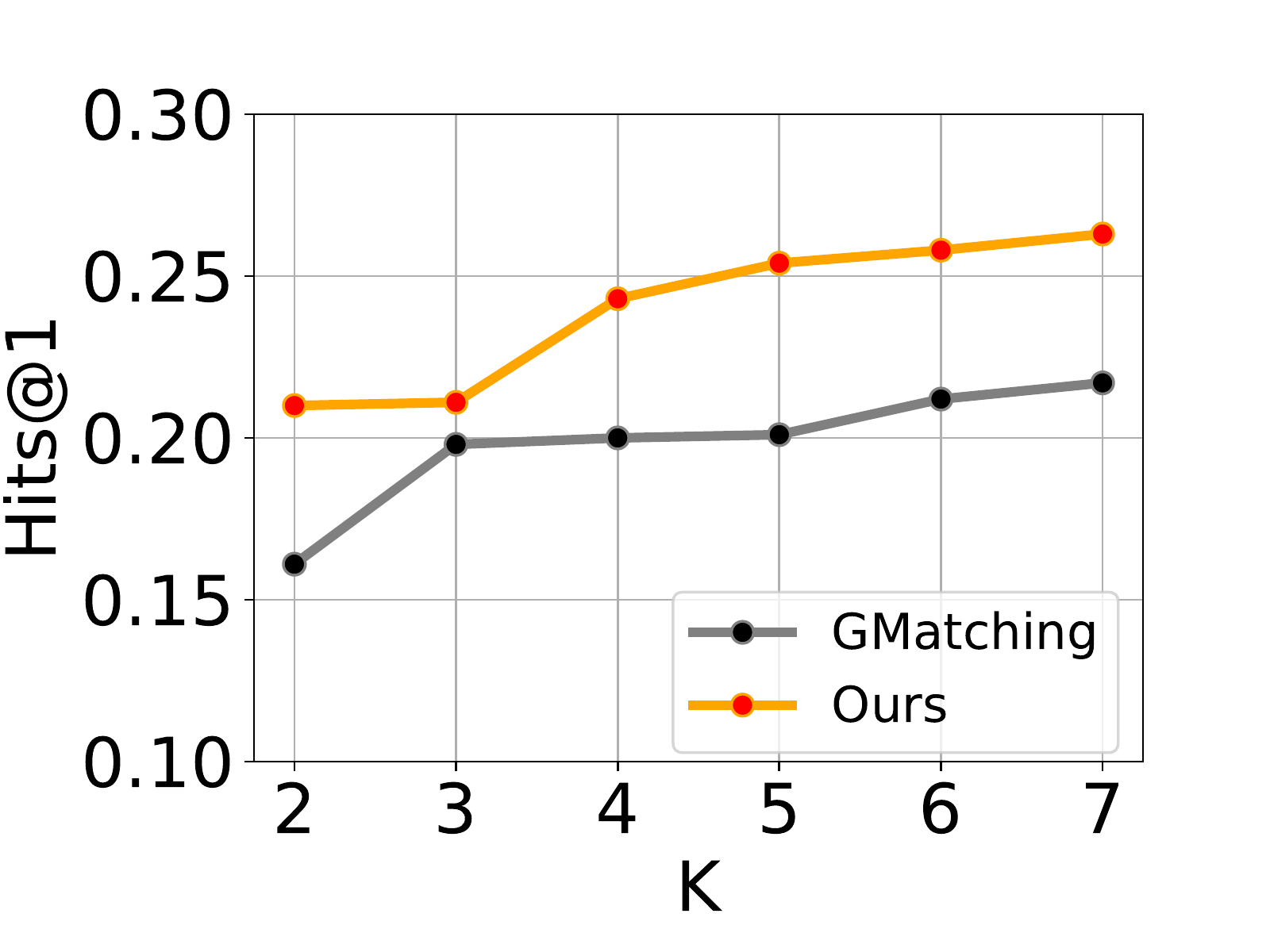}
\includegraphics[scale=0.255]{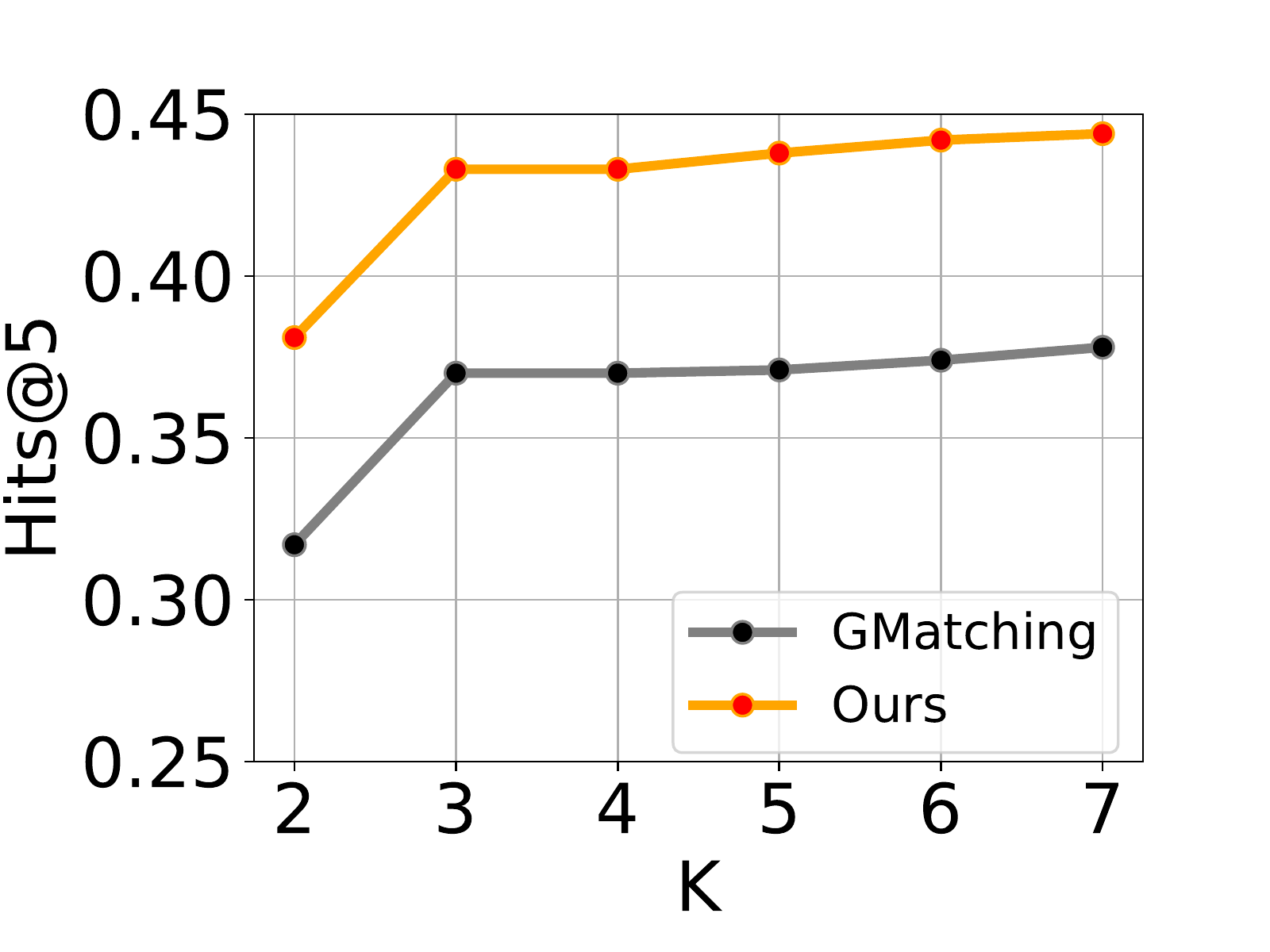}
\includegraphics[scale=0.255]{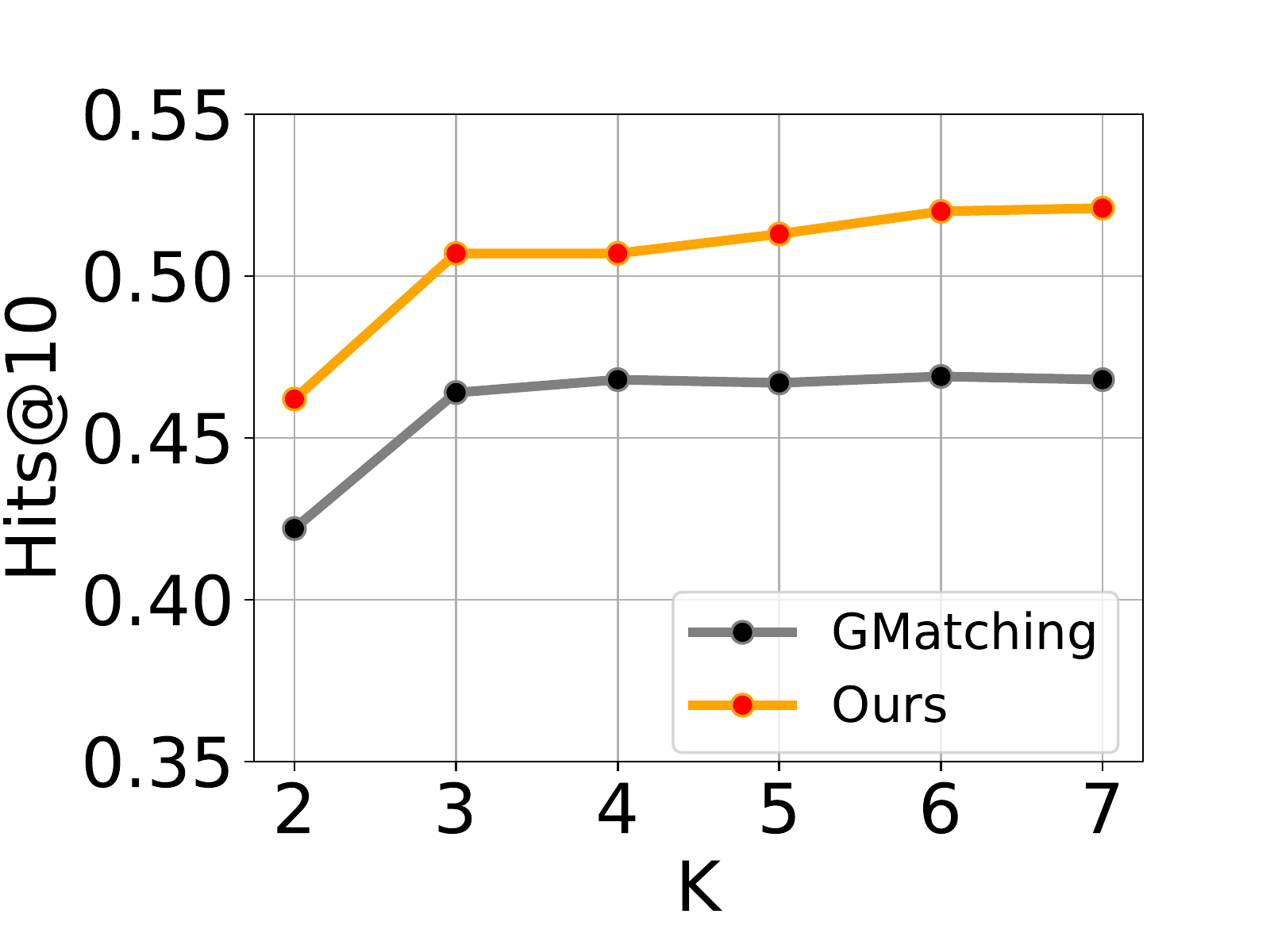}
\includegraphics[scale=0.255]{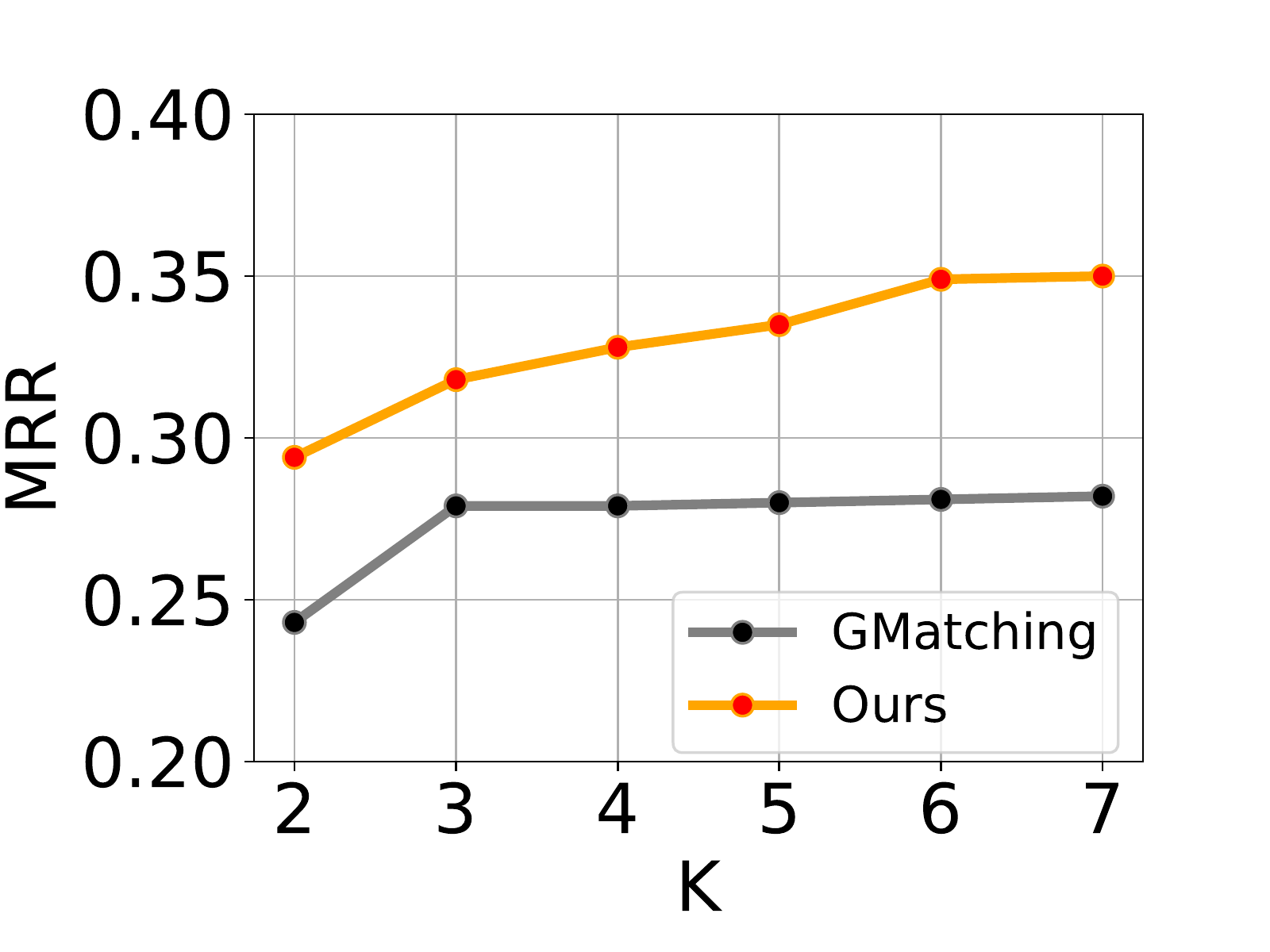}
\caption{Impact of few-shot size K. Our model consistently outperforms GMatching.} 
\label{fig: impact_shot_number}
\end{center}
\end{figure}

This work studies few-shot relation learning in KGs, thus we conduct experiment to analyze the impact of few-shot size $K$. Figure \ref{fig: impact_shot_number} reports the performances of our model and GMatching (MaxP) in NELL test data with different settings of $K$. According to the figure:
\begin{itemize}[leftmargin=*]
\item With the increment of $K$, performances of both models increase. It indicates that larger reference set can produce better reference set embedding for the relation.  
\item Our model consistently outperforms GMatching in different $K$, demonstrating the stability of the proposed model for few-shot relation completion in KGs. 
\end{itemize}

\subsubsection{Embedding Visualization}
\begin{figure}[hbt!]
\begin{center}
\includegraphics[scale=0.25]{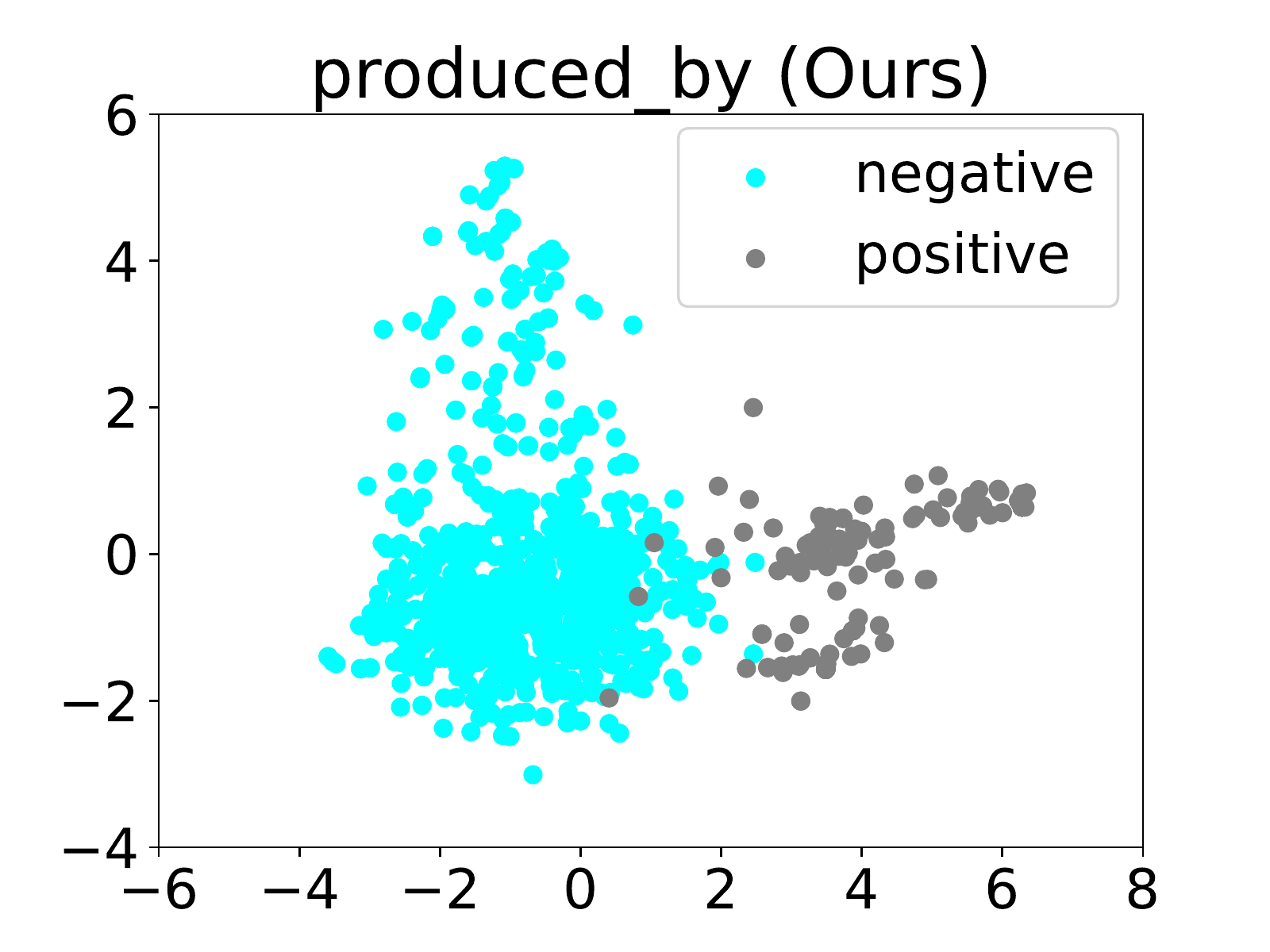}
\includegraphics[scale=0.25]{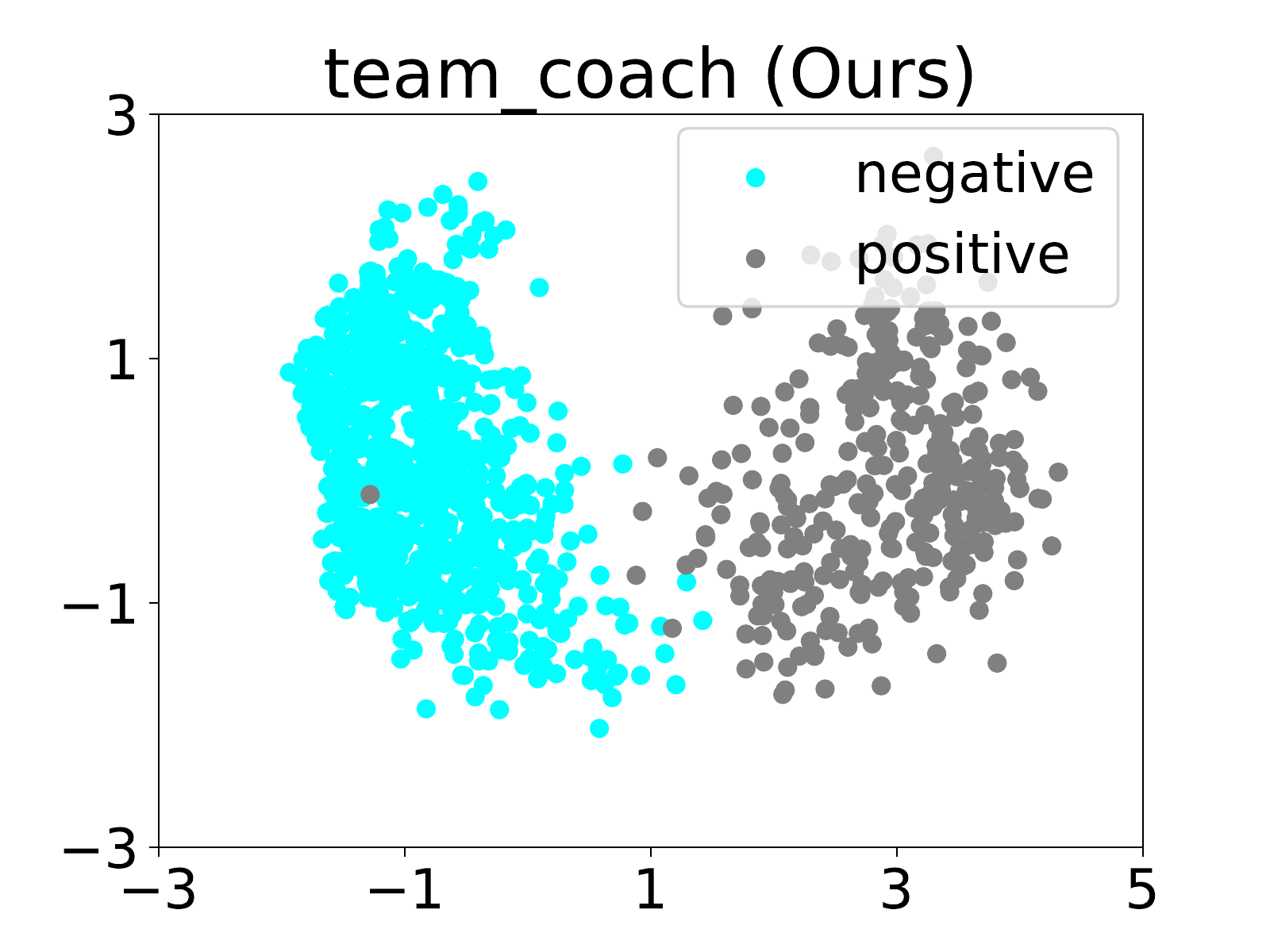}
\includegraphics[scale=0.25]{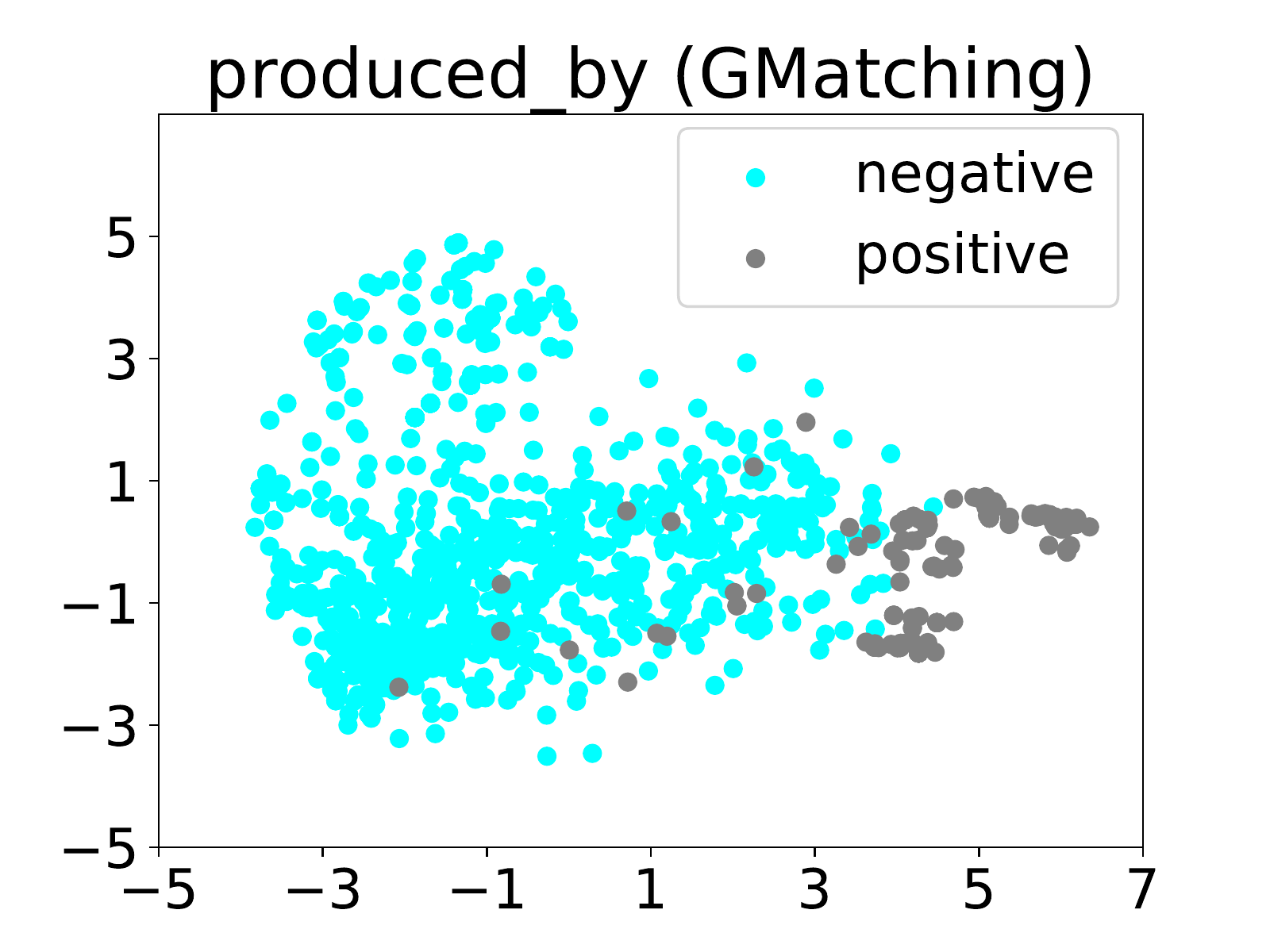}
\includegraphics[scale=0.25]{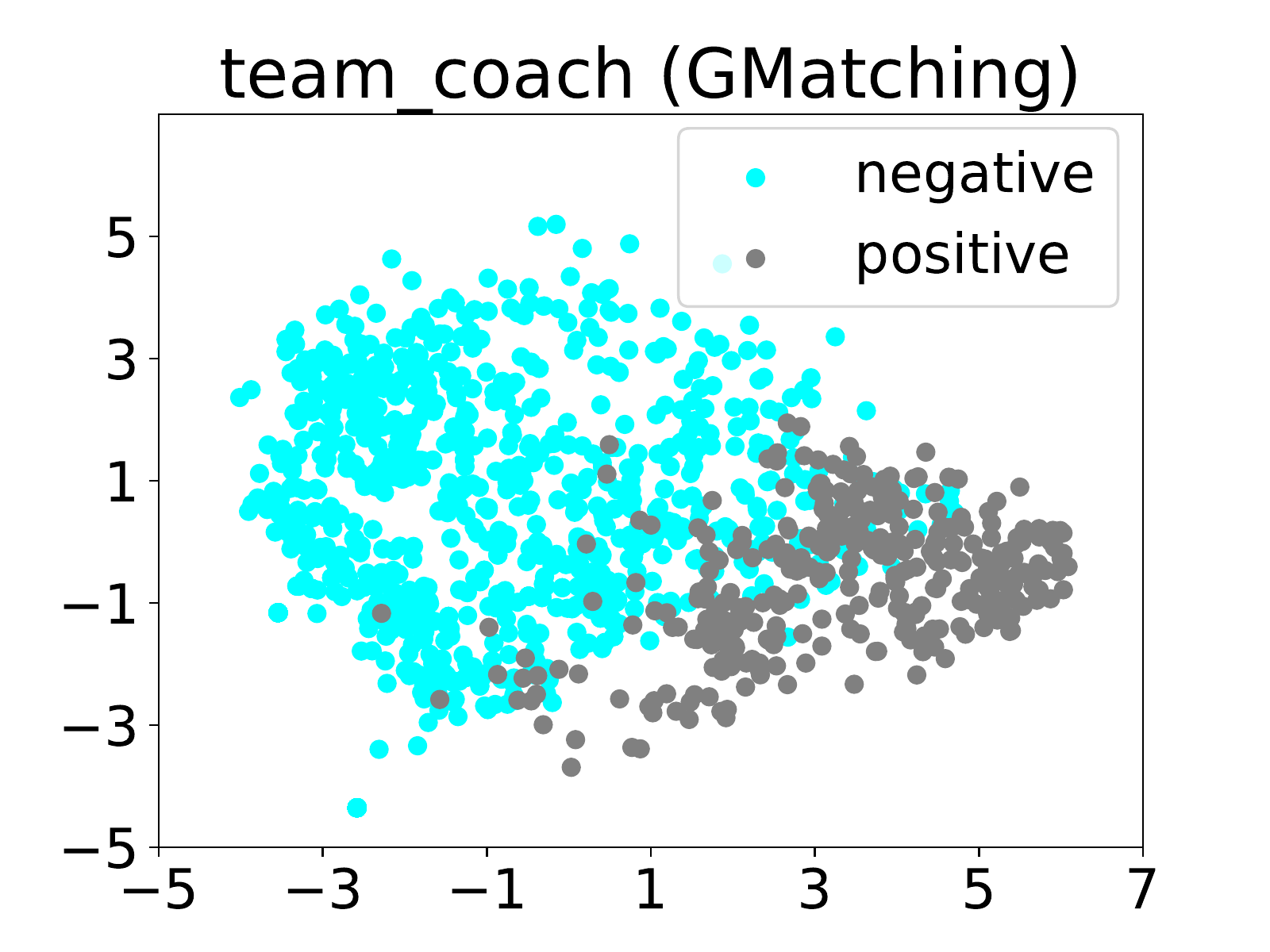}
\caption{Embedding visualization of positive and negative candidates of two selected relations. Our model can clearly discriminate embeddings of these two types of candidates.} 
\label{fig: visualization}
\end{center}
\end{figure}

To show a better performance comparison between our model and GMatching, we visualize the 2D embeddings of positive and negative candidate entity pairs for each relation. Figure \ref{fig: visualization} shows the visualization results of our model and GMatching for two test relations of NELL data, i.e., ``produced\_by'' and ``team\_coach'', which vary from each other in semantic meaning and size of positive/negative candidate set. According to the figure, both methods can distinguish embeddings of positive and negative candidates well. However, it is clear that our model achieves better performance and embeddings of two classes are clearly discriminated from each other, which further demonstrates the superior performance of our model in terms of visualization.

\section{Conclusion}
In this paper, we presented a new few-shot KG completion problem and proposed an innovative few-shot relation learning model, i.e., FSRL, to solve the problem. FSRL performs joint optimization of relation-aware heterogeneous neighbor encoder, recurrent autoencoder aggregation network and matching network. The extensive experiments on two public datasets demonstrate that FSRL can outperform state-of-the-art baseline methods. In addition, the ablation studies verify the effectiveness of each model component. As a new research problem, there are many opportunities for the next steps. The future work might consider utilizing a better model training process such as model-agnostic meta-learning or incorporating contextual information such as entity attributes or text description to improve the quality of entity embeddings.

\section*{Acknowledgments}
This work was supported by the Army Research Laboratory under Cooperative Agreement Number W911NF-09-2-0053 and the National Science Foundation awards \#1925607, \#1629914, \#1652525, \#1618448 and \#1849816.

\bibliographystyle{aaai}
\bibliography{reference}

\end{document}